\documentclass[10pt,twocolumn,letterpaper]{article}

\usepackage{wacv}

\usepackage{booktabs}
\usepackage{multirow}
\usepackage{subcaption}
\usepackage{amsmath,amssymb}

\usepackage[accsupp]{axessibility}

\definecolor{wacvblue}{rgb}{0.21,0.49,0.74}
\usepackage[pagebackref,breaklinks,colorlinks,allcolors=wacvblue]{hyperref}

\def\wacvPaperID{15}
\def\confName{WACV}
\def\confYear{2027}

\title{Illusion of Explainability: A Sanity Check for Spectral Band\\
Attribution in Satellite Time-Series Crop Classification}

\author{Sumit Tahiliani\\
SatSure Analytics India Pvt.\ Ltd.\\
{\tt\small sumit@satsure.co}
\and
Avi Mathur\\
Shiv Nadar University\\
{\tt\small avi.7mathur@gmail.com}
\and
Sandarbh Agrawal\\
SatSure Analytics India Pvt.\ Ltd.\\
{\tt\small sandarbh@satsure.co}
\and
Rakshith Sathish\\
SatSure Analytics India Pvt.\ Ltd.\\
{\tt\small rakshith@satsure.co}
\and
Ashwathi Nambiar\\
SatSure Analytics India Pvt.\ Ltd.\\
{\tt\small ashwathi@satsure.co}
}

\begin{document}
\maketitle
\begin{abstract}
Attribution methods are frequently validated by two informal criteria: convergence across methods and consistency with domain knowledge. Neither establishes that the identified features are the ones a task requires. The gap widens in multispectral remote sensing, where spectral bands overlap heavily in what they measure: a band can receive high attribution and still be substitutable because the signal that separates the classes survives elsewhere in the spectrum. We examine this on crop-type classification, evaluating GradientSHAP, Shapley sampling, and Occlusion on a U-TAE segmentation model trained on Sentinel-2 time series from the PASTIS benchmark. The attributions behave as the standard criteria would require: all three methods concentrate their aggregate rankings on vegetation-sensitive red, red-edge, near-infrared, and SWIR regions, consistent with established remote sensing findings, and the Shapley and Occlusion rankings are order-sensitive on the frozen model, producing deletion and insertion curves that are sharp relative to a random ordering and invert under reversal. Neither property predicts retraining behavior. Aggregate consensus conceals per-class disagreement, with mean Spearman correlation between method pairs ranging from 0.09 to 0.79 and varying under de-correlation. In the retraining experiment, models trained from scratch on the five lowest-ranked bands perform as well as or better than those trained on the five highest, whereas the ranking of a second, equally accurate training run with identical configuration and seed selects a top five that retrains better. A redundancy analysis provides a plausible explanation: across all 252 five-band subsets, effective dimensionality remains between 1.35 and 2.62, suggesting that fine-grained band-level selection cannot be reliably inferred from attribution rankings in this input space. Attribution rankings that look plausible and faithful therefore need not identify bands whose selection improves retrained performance, particularly when spectral information is redundant.
\end{abstract}

\section{Introduction}
\label{sec:intro}

Crop-type classification is formulated as a semantic segmentation problem with the objective of assigning a crop label to each pixel of a satellite image time series. Many crops are spectrally indistinguishable on any single date and separable only by their phenological trajectories over the season, making multi-temporal, multispectral input essential. At scale, we rely on deep neural networks to achieve this; however, this comes at a cost to interpretability, since these outputs feed subsidy verification, yield forecasting, and decisions about which bands future sensors should carry. 

\begin{figure}[t]
\centering
\includegraphics[width=\columnwidth]{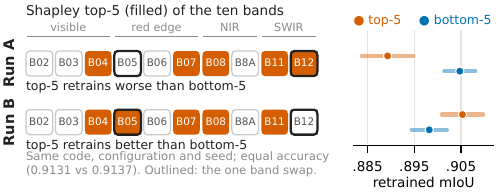}
\caption{Two runs of one training configuration (same code and seed, equal accuracy) give Shapley rankings that differ by a single band swap (outlined). Both show order-sensitive frozen-model behaviour under the evaluated checks, yet retraining on the top-5 is significantly worse than on the bottom-5 for Run~A and significantly better for Run~B (mean and 95\% CI over five seeds). Orange: top-5; blue: bottom-5.}
\label{fig:overview}
\end{figure}

The standard way to recover interpretability is through post hoc attribution. SHAP unifies a family of additive attribution methods under a Shapley-value framework \cite{lundberg2017}; GradientSHAP approximates these values via expected gradients over a baseline distribution, building on Integrated Gradients \cite{sundararajan2017,erion2021}; and occlusion-based attribution instead measures the change in output when a feature is masked \cite{zeiler2014}, closely related to perturbation-based faithfulness metrics such as insertion and deletion curves \cite{samek2017,petsiuk2018}. In remote sensing, these scores are aggregated into per-band rankings and checked against domain knowledge, with the red, red-edge, and near-infrared regions well established as the most discriminative for vegetation and crop mapping \cite{immitzer2016,vuolo2018}. Recovering these bands, especially across several methods, is commonly taken as validation. 

Neither check supports that conclusion. Agreement with a domain prior indicates that an attribution is plausible, not faithful, and masking-based faithfulness describes a frozen model's response to counterfactual inputs, not a band's information content for the task. Hooker et al.\ \cite{hooker2019} made this concrete with ROAR, showing that in large-scale image classification, many pixel-attribution methods are no better than random ranking when accuracy is measured by retraining on data with the top-ranked features removed rather than by masking a frozen model. Sanity checks in a similar spirit have shown that saliency maps can be insensitive to model and label randomisation \cite{adebayo2018}, and that ROAR's own retraining protocol leaks information through the masking pattern \cite{rong2022}.

Our protocol adapts ROAR to spectral bands rather than pixels, and to multi-temporal multi-class segmentation rather than single-label classification. The gap between plausibility and necessity is especially sharp in multispectral data, where adjacent bands are correlated by construction. Shapley-value attributions may behave unreliably under feature dependence: the coalition semantics become ambiguous \cite{sundararajan2020}, off-manifold baselines assign mass to correlated features that carry no independent signal \cite{kumar2020,strobl2008}, and accurate estimation requires modelling the dependence structure explicitly \cite{aas2021}. A model may therefore attribute heavily to one band while the same discriminative signal remains recoverable from a correlated subset. 

Therefore, the ranking describes the solution this model happened to find, and does not by itself establish which bands are uniquely required by the task. To quantify how much independent signal a band subset actually carries, we use the participation ratio, a rotation-invariant summary of a covariance spectrum, which is used in neuroscience to measure the effective dimensionality of correlated population signals \cite{gao2015}. We test this on PASTIS \cite{garnot2021} using a U-TAE segmentation model, evaluating GradientSHAP, Shapley sampling, and Occlusion at three levels of evidence: plausibility (agreement with domain knowledge and across methods), faithfulness (the frozen model's response to masking in ranked order), and selection utility (performance when the model is retrained on the selected bands). 

In the retraining experiment, models retrained from scratch on the five lowest-ranked bands perform as well as or better than those trained on the five highest-ranked bands, whereas the ranking of a second, equally accurate training run with identical configuration and seed selects a top five that retrains better (\cref{fig:overview}). Thus, the rankings pass the first two evaluation criteria while failing to distinguish the subsets under retraining, and the redundancy analysis provides a plausible explanation. Our contribution is diagnostic: we propose no new attribution method, and we do not claim the aggregate vegetation-sensitive band finding is wrong. We show instead that the criteria used to validate an attribution can fail to distinguish an explanation that identifies uniquely important features from one that selects an arbitrary member of a redundant set. In correlated sensor settings, this distinction should therefore be tested rather than assumed.

Related effects are known elsewhere: equally accurate models can rely on different features~\cite{fisher2019,dong2020,damour2022}, explanation methods often disagree~\cite{krishna2024}, and top-$k$ truncation of an importance ranking ignores redundancy~\cite{guyon2003,peng2005}. We show how they combine in spectral band selection, where retraining exposes them.

\section{Experimental setup}
\label{sec:method}
\paragraph{Dataset.} We use PASTIS \cite{garnot2021}, a panoptic agricultural satellite image time-series benchmark comprising 2{,}433 patches of $128\times128$ pixels across four regions of France, each a Sentinel-2 optical time series with 10 spectral bands (B02, B03, B04, B05, B06, B07, B08, B8A, B11, B12; listed in the supplementary material) and between 38 and 61 acquisitions. The dataset defines 18 crop classes plus a background class, which we ignore (we call this the 18-class setting), with a heavy long-tailed class distribution - background and meadow alone cover roughly 68\% of labeled pixels.
Unless stated otherwise we use a fixed split of PASTIS's official folds: folds 1--3 for training (1{,}455 patches), fold 4 for validation (482) and fold 5 as the held-out test set (496). For the fold-rotation experiment, fold $f$ is the test set, fold $f{-}1$ (cyclically) validation and the remaining three folds training, so that $f{=}5$ recovers the fixed split. Cloud filtering is performed using a Sentinel-2 Scene Classification Layer (SCL)-based protocol; full details are given in the supplementary material.

\paragraph{Model.} We use U-TAE \cite{garnot2021} for crop classification in the PASTIS dataset. It is an architecture consisting of a U-Net encoder with a temporal self-attention bottleneck for satellite image time series.
For the highly imbalanced classes we used focal loss instead of cross-entropy, which down-weights well-classified pixels, AdamW instead of Adam, a linear warmup plus cosine learning-rate schedule instead of a flat rate, and gradient-norm clipping at 0.5.
The attribution model is trained on all ten bands for 91 epochs; every retrained model in Section~\ref{sec:res-retrain} is trained for 50 epochs with the same schedule, so arms are compared at a matched training budget. We train the attribution model twice, as Run~A and Run~B, with identical code, configuration, random seed (42) and software environment. cuDNN is set to deterministic mode, but other GPU operations are not deterministic, so the two runs diverge during training and reach different solutions of equal accuracy (test mIoU 0.9131 and 0.9137). They are two equally accurate solutions under nominally identical settings, used as a case study rather than as seed replicates; independently seeded models are analysed separately. GradientSHAP was computed for Run~B only, so attributions, cross-method agreement and faithfulness use Run~B; Run~A's Shapley and Occlusion rankings are also evaluated for faithfulness, and Run~A's Shapley ranking defines the main retraining arms. Comparing the two runs measures ranking stability and retraining utility.
Background, and in the reduced label space every non-selected crop class, is mapped to an ignore label that is excluded from the loss and from all metrics, so metrics are computed over crop classes only.

\paragraph{Attribution methods.} We transform the model's class-specific spatio-temporal predictions into a spectral importance vector containing one score for each of the ten input bands for each of the seven classes being predicted. We compute this band-level importance with three methods, each producing one score per class and band on the held-out test set: GradientSHAP sums attributions over timesteps, while Occlusion and Shapley mask each band across all timesteps.

\begin{itemize}
\item \textbf{GradientSHAP.} We implement GradientSHAP following the expected-gradients formulation of \cite{erion2021}, using baselines drawn from a background data distribution rather than a single fixed reference.
References are real per-pixel time series from the training folds, and absolute attributions, averaged over 50 Monte Carlo samples per patch and class, are summed over timesteps (settings in the supplementary material).

\item \textbf{Shapley.} Shapley evaluates the importance of bands by calculating their average marginal contribution across different subsets (referred to as coalitions) of all the available bands. For each input band, the contribution is determined by the change in model output when the band is added to a subset compared to the same subset without the said band.
We estimate these average marginal contributions with the Monte-Carlo permutation estimator~\cite{strumbelj2010} using antithetic pairs~\cite{mitchell2022}: 30 band orderings per patch (15 and their reverses), adding bands one at a time to an all-zero input and recording each band's change in the class probability, so the values are sampled rather than exact (details in the supplementary material). Bands are ranked by their signed Shapley value averaged over patches and classes.

\item \textbf{Occlusion.}
Occlusion is a leave-one-out perturbation: each band in turn is set to zero across all timesteps, and the drop in the class probability over the class's ground-truth pixels is averaged across patches, giving one score per class and band.
\end{itemize}

\paragraph{Cross-method agreement.} For each class, we compare the resulting band-ranking vectors across attribution methods using Spearman rank correlation.

\paragraph{Faithfulness protocol.} We evaluate each ranking with deletion and insertion curves \cite{samek2017,petsiuk2018}: for deletion, we start from all 10 bands and remove them one at a time in ranked order, plotting class probability at each step (a faithful ranking should show a steep early drop); for insertion, we start from zero bands and add them in ranked order (a faithful ranking should show a steep early rise). We summarize each curve by its area under the curve (AUC), and compare against a random-order control. Bands are masked by replacing their normalised reflectance values with zero at every timestep and every pixel of the patch, so a masked band is absent from the entire time series.
Because zero-filling is not a natural observation and may measure the model's reaction to an off-manifold input rather than band importance~\cite{hooker2021unrestricted}, we repeat the evaluation with two further baselines: the per-band dataset mean, and a \emph{marginal} baseline that substitutes the same band from a randomly drawn other patch, preserving its distribution and temporal profile while breaking its dependence on the retained bands.

\paragraph{Retraining and identifiability evaluation.} We assess operational band-selection utility by testing whether retraining on the top-ranked subset consistently outperforms lower-ranked and random subsets under the same architecture, dataset and training protocol. Following ROAR \cite{hooker2019}, we additionally \emph{retrain} U-TAE from scratch on band subsets selected by rank, using the full temporal sequence but only the selected spectral channels, rather than only masking a frozen model, which avoids the mask-pattern leakage of pixel-level ROAR~\cite{rong2022}; every compared top- and bottom-5 holds two 10\,m bands, so spatial resolution does not separate them. We use Shapley rankings for this step because Shapley showed the strongest agreement with Occlusion in the cross-method analysis; the main arms use Run~A's ranking (top-5 B04 B07 B11 B08 B12).
We retrain the top-, bottom- and random five-band subsets, top- and bottom-$k$ subsets for $k\in\{1,2,3\}$, and the full ten bands on the fixed split, with five training seeds per five-band arm and three for the smaller subsets. Because seed and fold variance are different noise sources~\cite{bouthillier2021}, we also rotate the test fold, and we retrain twenty further random five-band subsets to obtain the distribution of performance across subsets.
Rankings are computed on the test fold, which also evaluates fixed-split retraining; this can only favour the top-ranked subsets. Test folds 1--4 of the fold rotation share no patches with the ranking data, and recomputing rankings on the validation fold for Run~B and the two seeded models gives near-identical orders ($\rho=0.89$--$0.99$) with Run~B's top- and bottom-5 unchanged, so its retraining comparison involves no test-set reuse. The comparison is run in both label spaces, using the full 18-class model's own Shapley ranking for the 18-class setting (three seeds per arm), and with the top- and bottom-5 of Run~B's ranking (five seeds per arm).

\paragraph{Statistical analysis.} Seed-level comparisons use Welch's $t$-test and an exact permutation test, and fold-level comparisons are paired by fold (paired $t$-test and exact sign-flip test).
We test equivalence with two one-sided tests (TOST) against a margin of twice the pooled seed SD (0.008 mIoU), the run-to-run noise floor. $p$-values are Holm-corrected over the six pairwise seed comparisons; the Run~B comparison is a separate follow-up test.

\paragraph{Spectral redundancy analysis.} For a $k$-band subset, we compute the participation ratio $\mathrm{PR} = (\sum_i \lambda_i)^2 / \sum_i \lambda_i^2$ over the eigenvalues $\lambda_i$ of the $k\times k$ Pearson correlation matrix of the selected bands \cite{gao2015}, and a variance coverage $\mathrm{VC}=\sum_j v_j R_j^2$, where $R_j^2$ measures how well the subset reconstructs the $j$-th principal component of all 10 bands and $v_j$ is its variance share.
We compute both for every subset of every size $k=2,\dots,9$, on pixels from cloud-free timesteps. Since both measures are linear, we add two nonlinear ones: a \emph{nonlinear VC}, in which the principal components are reconstructed from the subset with gradient-boosted trees on held-out pixels, and a \emph{task probe}, a gradient-boosted pixel classifier trained on each subset's cloud-free temporal profiles and scored by macro-F1 on the test fold.

\section{Results and discussion}
\label{sec:results}

\paragraph{Baseline segmentation performance and class selection.} We initially evaluate U-TAE using all 18 crop classes of the PASTIS dataset. The results showed substantially different predictive performance across classes, motivating a reduced-class experiment suite for the subsequent explainability experiments.
For diagnostic stability, we selected seven well-classified crop classes of the full-class model, to avoid poorly supported classes in the attribution analysis; on the validation fold they are seven of the eight classes with the highest F1 (the eighth is Beet). The classes shortlisted were: Winter rapeseed, Corn, Soft winter wheat, Meadow, Soy, Grapevine and Winter barley. All remaining classes were relabeled to background and a new model was trained using this modified label space.
Because background is excluded from the loss and from all metrics, the reduced task is seven-way discrimination among crop pixels. Because selecting well-classified classes favours well-separated ones, we treat the seven-class and full 18-class tasks as two settings: the seven-class setting is used for the attribution diagnostics, and the retraining comparison is run in both.

\begin{table}[t]
\caption{Performance of our U-TAE reproduction on PASTIS. The 18-class setting uses all 18 crop classes and all 10 spectral bands. The 7-class setting uses the selected seven crop classes with all remaining classes relabelled as background. Metrics are computed over crop classes. $^\dagger$Not reported; see the fixed-split row. $^\ddagger$Includes background in mIoU, so not directly comparable.}
\label{tab:baseline}
\centering\small
\setlength{\tabcolsep}{3pt}
\resizebox{\columnwidth}{!}{%
\begin{tabular}{lccc}
\toprule
Model & mIoU & Macro F1 & Weighted F1 \\
\midrule
Garnot \& Landrieu (5-fold)$^\ddagger$ & .631 & -- & -- \\
Ours (18-cl., 5-fold) & $.680 \pm .010$ & --$^\dagger$ & $.897 \pm .004$ \\
Ours (18-cl., fixed split) & $.674 \pm .012$ & $.789 \pm .009$ & $.890 \pm .006$ \\
Ours (7-class) & .913 & .954 & .967 \\
\bottomrule
\end{tabular}}
\end{table}

The 7-class model's performance is reported in \cref{tab:baseline}; all subsequent results use the 7-class setting unless stated otherwise, and per-class results for both settings are in the supplementary material.

\begin{figure*}[t]
\centering
\includegraphics[width=\textwidth]{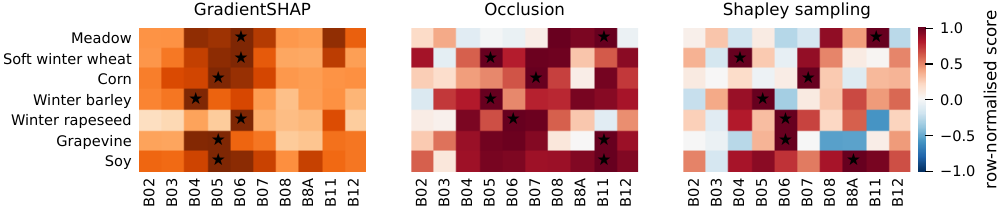}
\caption{Per-class band importance for the 7-class model from GradientSHAP (left), Occlusion (centre), and Shapley sampling values (right), each row normalised to its largest magnitude; all three are computed on Run~B. The methods exhibit different class-specific importance patterns despite their aggregate agreement. A star marks the top band for each crop class.}
\label{fig:perclass-attribution}
\end{figure*}

\paragraph{Aggregate attribution and domain knowledge.} At the aggregate level, all three methods assign substantial importance to the Near-Infrared (NIR, B08), Red Edge (B05, B06, B07), and Short-Wave Infrared (SWIR, B11) spectral regions. The top-ranked bands for each attribution method are summarized in \cref{tab:aggregate-rankings}. This alignment with established remote sensing literature looks, in isolation, like validated, method-agnostic evidence for band importance. However, this aggregate view can obscure substantial, setting-dependent disagreement at the per-class level.

\begin{table}[t]
\caption{Top-5 bands per attribution method for the 7-class model (Run~B), ranked by descending importance.}
\label{tab:aggregate-rankings}
\centering\small
\setlength{\tabcolsep}{4pt}
\begin{tabular}{lccccc}
\toprule
Method & 1st & 2nd & 3rd & 4th & 5th \\
\midrule
GradientSHAP & B06 & B04 & B05 & B11 & B07 \\
Occlusion & B07 & B11 & B05 & B08 & B06 \\
Shapley sampling & B07 & B11 & B08 & B04 & B05 \\
\bottomrule
\end{tabular}
\end{table}

\paragraph{Per-class attribution disagreement.}
At the per-class level, however, the bands identified as important vary substantially across crop classes and between methods (\cref{fig:perclass-attribution}), so aggregation masks disagreement. Similar method-dependent disagreement persists in the full 18-class setting, indicating that the phenomenon is not specific to the reduced label space (heatmaps in the supplementary material).
The global ranking is itself sensitive to how per-class scores are combined: averaging signed Shapley values over classes gives B07 as the top band, averaging their magnitudes gives B08, per-class normalisation gives B04, and pooling over all patches gives B08.

\paragraph{Cross-method agreement.} To quantify this disagreement, we computed the pairwise Spearman rank correlation ($\rho$) on the spectral importance feature vectors across classes.
Rank correlation compares orderings without requiring comparable magnitudes. The resulting mean rank correlations and top-band agreement rates are reported in \cref{tab:agreement}; per-class values are in the supplementary material.

\begin{table}[t]
\caption{Mean per-class Spearman rank correlation between attribution-method band rankings, and fraction of classes where compared methods agree on the single top band, for the 7-class setting. All rankings from Run~B (for PCA-10, Run~B of the PCA model).}
\label{tab:agreement}
\centering\small
\setlength{\tabcolsep}{4pt}
\resizebox{\columnwidth}{!}{%
\begin{tabular}{llcc}
\toprule
Input & Method pair & Mean $\rho$ & Top match \\
\midrule
Raw & GradientSHAP vs.\ Occlusion & 0.16 & 1/7 \\
Raw & Shapley vs.\ Occlusion & \textbf{0.78} & 4/7 \\
Raw & GradientSHAP vs.\ Shapley & 0.09 & 1/7 \\
PCA-10 & GradientSHAP vs.\ Occlusion & 0.61 & 4/7 \\
PCA-10 & Shapley vs.\ Occlusion & \textbf{0.79} & 6/7 \\
PCA-10 & GradientSHAP vs.\ Shapley & 0.57 & 3/7 \\
\bottomrule
\end{tabular}}
\end{table}

Agreement is neither uniformly high nor method-invariant: mean $\rho$ ranges from 0.09 to 0.79, and Shapley agrees with Occlusion far more than GradientSHAP does (0.78 vs.\ 0.16 on raw bands). Agreement is also setting-dependent: decorrelating the bands with PCA raises GradientSHAP's agreement with occlusion from 0.16 to 0.61 and with Shapley from 0.09 to 0.57, while Shapley--occlusion agreement is unchanged (0.78 to 0.79). Band correlation, rather than the methods themselves, accounts for most of the disagreement. Aggregate, class-averaged agreement statistics should therefore be reported with this instability made explicit.

\begin{figure*}[t]
\centering
\includegraphics[width=\textwidth]{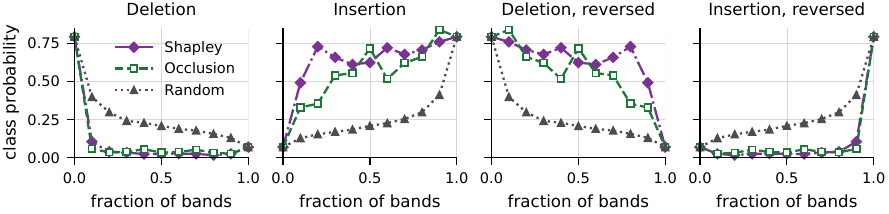}
\caption{Deletion and insertion curves for the 7-class model (Run~B), in the ranked order (left) and with the order reversed so the least important band is removed or inserted first (right). The reversed curves invert relative to the standard ordering, confirming order-sensitivity on the frozen model.}
\label{fig:faithfulness}
\end{figure*}

\paragraph{Faithfulness under perturbation.} The resulting AUC values are reported in \cref{tab:faithfulness}, while the corresponding insertion and deletion curves are shown in \cref{fig:faithfulness}.
Shapley and Occlusion outperform the random ordering on both deletion and insertion, while GradientSHAP does not outperform its random control under zero-filling. Run~A's Shapley and Occlusion rankings, which define the main retraining arms, pass the same checks (\cref{tab:faithfulness}). These results show that the rankings can be faithful and order-sensitive on a frozen model without establishing that the identified bands are uniquely or reliably necessary after retraining.

\begin{table}[t]
\caption{Deletion ($\downarrow$) and insertion ($\uparrow$) AUC for the 7-class model under three removal baselines: zero-filling, the per-band mean, and a marginal sample (the band taken from another patch). Each run is compared with its own random control.}
\label{tab:faithfulness}
\centering\small
\setlength{\tabcolsep}{2.6pt}
\begin{tabular}{lcccccc}
\toprule
& \multicolumn{2}{c}{Zero} & \multicolumn{2}{c}{Mean} & \multicolumn{2}{c}{Marginal} \\
\cmidrule(lr){2-3}\cmidrule(lr){4-5}\cmidrule(lr){6-7}
Ranking & Del & Ins & Del & Ins & Del & Ins \\
\midrule
\multicolumn{7}{l}{\textit{Run B}} \\
Shapley & \textbf{.074} & \textbf{.641} & \textbf{.225} & \textbf{.472} & .250 & .386 \\
Occlusion & .079 & .556 & .226 & .458 & \textbf{.242} & \textbf{.411} \\
GradientSHAP & .260 & .176 & .299 & .330 & .286 & .324 \\
Random & .245 & .246 & .343 & .352 & .324 & .326 \\
\midrule
\multicolumn{7}{l}{\textit{Run A (ranking used for the main retraining arms)}} \\
Shapley & \textbf{.076} & \textbf{.635} & \textbf{.233} & \textbf{.481} & \textbf{.253} & \textbf{.397} \\
Occlusion & .078 & .566 & .252 & .447 & .256 & .389 \\
Random & .259 & .259 & .341 & .346 & .326 & .327 \\
\bottomrule
\end{tabular}
\end{table}

\paragraph{Removal baseline.} Zero-filling creates spectral vectors the model never saw during training, so the curves may partly measure its reaction to off-manifold input~\cite{hooker2021unrestricted}. Shapley and Occlusion themselves use zero-filling, so zero-fill curves partly favour them; the per-band mean and a marginal sample are stricter tests. Shapley and Occlusion still outperform the random ordering under both (\cref{tab:faithfulness}), but Shapley's deletion-AUC gain over random falls from 0.171 to 0.118 and 0.074, and under the marginal baseline Occlusion overtakes Shapley. The ordering remains informative under these baselines, although its advantage varies. Reversing the band order (least-important first) makes deletion AUC markedly higher and insertion AUC markedly lower (\cref{fig:faithfulness}, right), so the faithfulness result depends on the attribution-derived ordering rather than on noise.

\begin{figure*}[t]
\centering
\includegraphics[width=\textwidth]{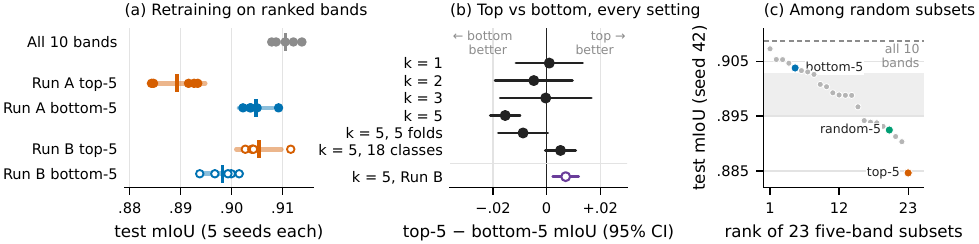}
\caption{Retraining U-TAE on ranked band subsets (7-class unless noted). (a) Test mIoU over five seeds per arm (dots; hollow for Run~B), with mean (tick) and 95\% CI (bar), for subsets chosen by Run~A's and Run~B's Shapley rankings. (b) Top-$k$ minus bottom-$k$ mIoU with 95\% CI; rows use Run~A's ranking except the 18-class row (that model's own) and the last (Run~B's). (c) Twenty random five-band subsets and the three named subsets at seed 42, ranked; the grey band is $\pm1$ seed SD around the median.}
\label{fig:retraining}
\end{figure*}

\paragraph{Retraining and identifiability.}
\label{sec:res-retrain}
Having evaluated plausibility and faithfulness, we next test selection utility after retraining. Across five random seeds per arm, the bottom-5 and random-5 subsets achieve similar retrained performance, and both outperform the top-5 subset (\cref{tab:retrain}, \cref{fig:retraining}a), despite their substantially different Shapley rankings. The top-5 deficit against bottom-5 is significant after Holm correction ($\Delta=-0.0155$ mIoU, exact permutation $p=0.008$, Holm-adjusted $p=0.048$); against random-5 it is borderline ($p=0.024$, Holm-adjusted $0.071$). Bottom-5 and random-5 do not differ significantly ($p=0.45$), although the experiment does not establish their equivalence either (TOST $p=0.053$; 90\% CI of the difference $-0.0027$ to $0.0081$ against a $\pm0.008$ margin). This indicates that, in this setting, the attribution ranking does not identify a uniquely necessary set of five bands, and if anything points away from the best subset. Importantly, redundancy does not make every subset equally informative: a single band loses about 0.11 mIoU, while several distinct five-band subsets retain comparable performance.

\begin{table}[t]
\caption{Retraining U-TAE on 7-class band subsets. Top-5 and Bottom-5 are the five highest and lowest ranked bands of Run~A's Shapley ranking; B03 is the single lowest-ranked band; Random-5 is an arbitrary five-band subset. Results are mean $\pm$ standard deviation across five random seeds (50 epochs). PCA: a single model trained for 91 epochs, as the attribution model (0.9131 on raw bands). $^\ast$Run~A's values ranked by magnitude, three seeds.}
\label{tab:retrain}
\centering\small
\setlength{\tabcolsep}{2.5pt}
\begin{tabular}{lccc}
\toprule
Band subset & mIoU & Macro F1 & Weighted F1 \\
\midrule
Full (\textit{all 10}) & .9106 $\pm$ .0024 & .9528 $\pm$ .0013 & .9662 $\pm$ .0016 \\
Top-5 & .8893 $\pm$ .0044 & .9407 $\pm$ .0026 & .9591 $\pm$ .0018 \\
Bottom-5 & .9047 $\pm$ .0027 & .9496 $\pm$ .0015 & .9634 $\pm$ .0013 \\
Random-5 & .9021 $\pm$ .0056 & .9481 $\pm$ .0031 & .9626 $\pm$ .0020 \\
Top-5$^\ast$ & .9014 $\pm$ .0023 & .9477 $\pm$ .0013 & .9627 $\pm$ .0010 \\
Bottom-5$^\ast$ & .8952 $\pm$ .0032 & .9443 $\pm$ .0018 & .9603 $\pm$ .0014 \\
B03 only & .8034 $\pm$ .0044 & .8874 $\pm$ .0029 & .9300 $\pm$ .0016 \\
PCA (\textit{10 comp.}) & .9144 & .9550 & .9675 \\
\bottomrule
\end{tabular}
\end{table}

\begin{figure}[t]
\centering
\includegraphics[width=0.85\columnwidth]{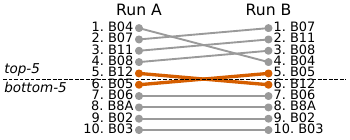}
\caption{Global Shapley rank of each band for Run~A and Run~B, which share configuration and seed and have the same accuracy. Highlighted bands cross the top-5 boundary.}
\label{fig:stability}
\end{figure}

\begin{figure*}[t]
\centering
\includegraphics[width=\textwidth]{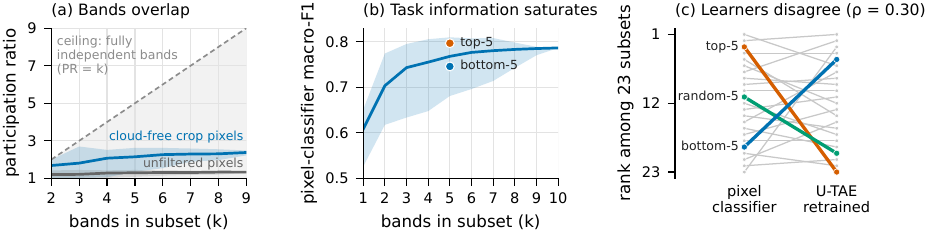}
\caption{(a) Participation ratio of every $k$-band subset (median and range) on cloud-free crop pixels and on unfiltered pixels; the dashed line is the ceiling for fully independent bands. (b) Macro-F1 of a pixel-level gradient-boosted classifier over all subsets of each size, with Run~A's Shapley top-5 and bottom-5 marked. (c) Rank of the 23 retrained five-band subsets under the pixel classifier and after U-TAE retraining.}
\label{fig:redundancy}
\end{figure*}

\paragraph{Distribution over five-band subsets.} We also retrained twenty more subsets drawn uniformly at random (\cref{fig:retraining}c). At the matched seed, Run~A's Shapley top-5 is the lowest of all 23 subsets (0.8847; next lowest 0.8903), and its five-seed mean confirms that this is not a single unlucky run. The spread across subsets is only modestly larger than seed-to-seed noise (SD 0.0058 vs.\ 0.0040), so most five-band subsets perform alike. Run~A's class-averaged Shapley values are all positive, so its bottom-5 holds the weakest, not harmful, bands. Ranking by magnitude instead swaps only B06 for B08, yet reverses the direction: that top-5 retrains 0.006 above its bottom-5 (three seeds, $p=0.06$; \cref{tab:retrain}), so the verdict depends on the aggregation as well as the training run. Subsets with and without B12 differ by less than 0.001 on average (B05: 0.003, $p=0.18$).

\paragraph{Fold-level variation.} Since fold variation is a separate noise source, we also rotated the test fold. Across folds, the all-band model varies with an SD of 0.0055, about twice its seed-to-seed variation. Bottom-5 outperforms top-5 on four of five folds and ties on the fifth (mean gap 0.0088, paired $t$-test $p=0.057$; \cref{fig:retraining}b), including three of the four test folds that share no patches with the ranking data. The direction of the top-5 deficit is thus consistent across test folds.

\paragraph{The 18-class setting.} To check that the result is not specific to seven well-separated classes, we repeated the comparison with all 18 classes, using the 18-class model's own ranking. Every five-band subset now loses 0.026--0.032 mIoU against all ten bands, and top-5 retrains slightly above bottom-5 and random-5 (by 0.005 and 0.006), although neither difference is significant (Welch $p=0.064$, $0.098$). The seven-class reversal therefore does not carry over, but neither does a clear benefit from following the ranking.

\paragraph{Subset size.} Finally, we retrained the top- and bottom-$k$ bands for $k=1,2,3$. At each size the two show no significant difference (all $|\Delta|\le0.005$, $p>0.4$), and random subsets perform alike (\cref{fig:retraining}b). Across the tested subset sizes, folds and label spaces, Run~A's ranking thus shows no consistent advantage for its top-ranked bands.

\paragraph{Stability across equally accurate models.} For band selection to be reliable, its practical recommendations should remain stable across comparably accurate models, so we compared the rankings of Run~A and Run~B. The global orders agree loosely (Shapley $\rho=0.92$, Occlusion $\rho=0.77$), but the parts practitioners act on do not (\cref{fig:stability}): the top Shapley band moves from B04 to B07, and B12 and B05 exchange places across the top-5 boundary. Two further all-band models trained with different seeds behave the same way: across all four models, only B04, B07 and B08 remain in the Shapley top-5 (supplementary material). A ranking thus partly reflects the solution a training run found and does not, by itself, establish which bands are most useful after retraining.

\paragraph{Ranking of an equally accurate model.} Since Run~A and Run~B rank the bands differently, we also retrained on the top-5 and bottom-5 of Run~B's ranking, which differ from Run~A's only by exchanging B12 and B05. The outcome reverses (\cref{fig:overview}): Run~B's top-5 retrains significantly \emph{better} than its bottom-5 ($0.9053\pm0.0036$ vs.\ $0.8982\pm0.0030$ over five seeds; Welch $p=0.010$), and is higher on every seed. The two top-5 sets share four bands, so the reversal turns on one choice: B05 rather than B12 adds 0.016 mIoU here and 0.007 in the bottom-5 pair, versus average presence effects of 0.003 (B05) and under 0.001 (B12) across random subsets. Bootstrapping test patches, which also absorbs per-patch Monte Carlo noise, gives 95\% CIs of B12$-$B05 that exclude zero with opposite signs in the two runs, suggesting that the swap reflects differences between the learned solutions rather than sampling noise in the attributions. Whether following the ranking helps or hurts thus depends on which of two equally accurate training runs was explained.

\paragraph{Spectral redundancy analysis.}
If all 5 bands are perfectly correlated, we would get PR = 1, and if completely uncorrelated, PR = 5.
Computed on cloud-free timesteps and crop pixels from the training folds, the Shapley-selected top-5 subset has PR $\approx 2.39$ and the bottom-5 subset PR $\approx 1.81$, both far below the five independent signals the bands could carry. Both subsets reconstruct most of the spectral variance of the full 10-band space (VC = 0.99 and 0.93, respectively). Across all $\binom{10}{5}=252$ five-band subsets, PR ranges from 1.35 to 2.62, and across every subset size it stays well below $k$ (\cref{fig:redundancy}a). Measured on unfiltered pixels, PR is lower still (1.08--1.45), because clouds make all bands co-vary.

Reprojecting all 10 bands into their principal components and retraining U-TAE on this space shows the same picture: in the 7-class setting, the PCA-10 model (0.9144) performs on par with the raw-band model trained the same way (0.9131).
The leading three principal components explain 98.4\% of spectral variance (loadings in the supplementary material), consistent with the ten bands spanning far fewer independent axes, a plausible explanation for why the top-5 versus bottom-5 distinction brings no retraining advantage.

\paragraph{Nonlinear and task-aware redundancy.} PR and VC are linear and may miss nonlinear dependence. Reconstructing the principal components with gradient-boosted trees raises five-band coverage from 0.74--0.99 to 0.85--0.99, so the bands are, if anything, more redundant than the linear measures suggest. To measure task information directly, we trained a pixel classifier on each subset's cloud-free temporal profiles; it saturates after about four bands (\cref{fig:redundancy}b) and, unlike U-TAE, it favours the Shapley top-5 (92nd percentile of 252 subsets; bottom-5 at the 19th). Across the 23 retrained subsets, none of these measures predicts U-TAE's accuracy (all $|\rho|\le0.30$, $p\ge0.16$; \cref{fig:redundancy}c). Thus, band necessity depends on the learner: bands that matter to a pixel classifier can be dispensable to a model that also exploits spatial context.

\paragraph{Implications for geospatial XAI.}
We do not claim that attribution methods are useless: the aggregate vegetation-sensitive band finding is not spurious. Rather, in redundant multispectral settings, masking-based faithfulness and agreement with domain priors are useful but insufficient validation criteria. Retraining on a few band subsets is a cheap additional check, but should use rankings from more than one trained model: as Run~A and Run~B show, the conclusion from one model's ranking can reverse for an equally accurate second model.
Effective dimensionality is a useful warning rather than a verdict: it signals that many band subsets are likely to suffice, but not which one retrains best, which retraining directly evaluates. In practice, report rankings under two aggregations and two training runs, use a non-zero masking baseline, and retrain a few subsets before selecting bands.

\paragraph{Conclusion.} Taken together, the results show that attribution rankings can agree with domain knowledge and provide faithful, order-sensitive explanations of a frozen model while failing to distinguish bands that are uniquely important from redundant substitutes after retraining.
With seven classes, Run~A's top-ranked bands retrain worse than its lowest-ranked; elsewhere the ranking shows no consistent advantage, and a second run's ranking or a magnitude aggregation reverses the result, so a retraining-based verdict depends on the model and on analysis choices. Spectral redundancy helps explain why many distinct five-band subsets perform alike. Attribution rankings should therefore not be used alone to justify spectral-band selection; in applications where the choice of bands matters, it is better to retrain and report ranking stability across runs.

\section{Limitations}
\label{sec:limitations}

We evaluate a single backbone (U-TAE) on a single dataset (PASTIS), so how far the findings depend on this architecture or on the spectral correlation of Sentinel-2 over French farmland is unknown. Retraining uses subsets chosen from the 10-band Shapley ranking, which need not transfer to a 5-band retrain.
The seven-class setting favours well-separated classes (the 18-class retraining checks this), and Run~A's ranking comes only from the test fold, which also evaluates its fixed-split retraining. Arms use at most five seeds (three for 18 classes), only faithfulness uses non-zero baselines, and temporal or spatial attribution is not studied.

{
    \small
    \bibliographystyle{ieeenat_fullname}
    \bibliography{main}

@article{immitzer2016,
  author  = {Immitzer, Markus and Vuolo, Francesco and Atzberger, Clement},
  title   = {First Experience with {Sentinel-2} Data for Crop and Tree Species Classifications in Central {Europe}},
  journal = {Remote Sensing},
  volume  = {8}, number = {3}, pages = {166}, year = {2016}
}

@article{vuolo2018,
  author  = {Vuolo, Francesco and Neuwirth, Martin and Immitzer, Markus and Atzberger, Clement and Ng, Wai-Tim},
  title   = {How Much Does Multi-Temporal {Sentinel-2} Data Improve Crop Type Classification?},
  journal = {International Journal of Applied Earth Observation and Geoinformation},
  volume  = {72}, pages = {122--130}, year = {2018}
}

@inproceedings{lundberg2017,
  author    = {Lundberg, Scott M. and Lee, Su-In},
  title     = {A Unified Approach to Interpreting Model Predictions},
  booktitle = {Advances in Neural Information Processing Systems (NeurIPS)},
  year      = {2017}
}

@article{erion2021,
  author  = {Erion, Gabriel and Janizek, Joseph D. and Sturmfels, Pascal and Lundberg, Scott M. and Lee, Su-In},
  title   = {Improving Performance of Deep Learning Models with Axiomatic Attribution Priors and Expected Gradients},
  journal = {Nature Machine Intelligence},
  volume  = {3}, pages = {620--631}, year = {2021}
}

@inproceedings{zeiler2014,
  author    = {Zeiler, Matthew D. and Fergus, Rob},
  title     = {Visualizing and Understanding Convolutional Networks},
  booktitle = {European Conference on Computer Vision (ECCV)},
  year      = {2014}
}

@inproceedings{garnot2021,
  author    = {Sainte Fare Garnot, Vivien and Landrieu, Loic},
  title     = {Panoptic Segmentation of Satellite Image Time Series with Convolutional Temporal Attention Networks},
  booktitle = {IEEE/CVF International Conference on Computer Vision (ICCV)},
  year      = {2021}
}

@inproceedings{sundararajan2017,
  author    = {Sundararajan, Mukund and Taly, Ankur and Yan, Qiqi},
  title     = {Axiomatic Attribution for Deep Networks},
  booktitle = {International Conference on Machine Learning (ICML)},
  year      = {2017}
}

@article{samek2017,
  author  = {Samek, Wojciech and Binder, Alexander and Montavon, Gr{\'e}goire and Lapuschkin, Sebastian and M{\"u}ller, Klaus-Robert},
  title   = {Evaluating the Visualization of What a Deep Neural Network Has Learned},
  journal = {IEEE Transactions on Neural Networks and Learning Systems},
  volume  = {28}, number = {11}, pages = {2660--2673}, year = {2017}
}

@inproceedings{petsiuk2018,
  author    = {Petsiuk, Vitali and Das, Abir and Saenko, Kate},
  title     = {{RISE}: Randomized Input Sampling for Explanation of Black-Box Models},
  booktitle = {British Machine Vision Conference (BMVC)},
  year      = {2018}
}

@inproceedings{hooker2019,
  author    = {Hooker, Sara and Erhan, Dumitru and Kindermans, Pieter-Jan and Kim, Been},
  title     = {A Benchmark for Interpretability Methods in Deep Neural Networks},
  booktitle = {Advances in Neural Information Processing Systems (NeurIPS)},
  year      = {2019}
}

@article{gao2015,
  author  = {Gao, Peiran and Ganguli, Surya},
  title   = {On Simplicity and Complexity in the Brave New World of Large-Scale Neuroscience},
  journal = {Current Opinion in Neurobiology},
  volume  = {32}, pages = {148--155}, year = {2015}
}

@inproceedings{adebayo2018,
  author    = {Adebayo, Julius and Gilmer, Justin and Muelly, Michael and Goodfellow, Ian and Hardt, Moritz and Kim, Been},
  title     = {Sanity Checks for Saliency Maps},
  booktitle = {Advances in Neural Information Processing Systems (NeurIPS)},
  year      = {2018}
}

@inproceedings{rong2022,
  author    = {Rong, Yao and Leemann, Tobias and Borisov, Vadim and Kasneci, Gjergji and Kasneci, Enkelejda},
  title     = {A Consistent and Efficient Evaluation Strategy for Attribution Methods},
  booktitle = {International Conference on Machine Learning (ICML)},
  pages     = {18770--18795}, year = {2022}
}

@inproceedings{sundararajan2020,
  author    = {Sundararajan, Mukund and Najmi, Amir},
  title     = {The Many {Shapley} Values for Model Explanation},
  booktitle = {International Conference on Machine Learning (ICML)},
  pages     = {9269--9278}, year = {2020}
}

@inproceedings{kumar2020,
  author    = {Kumar, I. Elizabeth and Venkatasubramanian, Suresh and Scheidegger, Carlos and Friedler, Sorelle A.},
  title     = {Problems with {Shapley}-Value-Based Explanations as Feature Importance Measures},
  booktitle = {International Conference on Machine Learning (ICML)},
  pages     = {5491--5500}, year = {2020}
}

@article{aas2021,
  author  = {Aas, Kjersti and Jullum, Martin and L{\o}land, Anders},
  title   = {Explaining Individual Predictions When Features Are Dependent: More Accurate Approximations to {Shapley} Values},
  journal = {Artificial Intelligence},
  volume  = {298}, pages = {103502}, year = {2021}
}

@article{strumbelj2010,
  author  = {{\v{S}}trumbelj, Erik and Kononenko, Igor},
  title   = {An Efficient Explanation of Individual Classifications Using Game Theory},
  journal = {Journal of Machine Learning Research},
  volume  = {11}, pages = {1--18}, year = {2010}
}

@inproceedings{bouthillier2021,
  author    = {Bouthillier, Xavier and Delaunay, Pierre and Bronzi, Mirko and Trofimov, Assya and Nichyporuk, Brennan and Szeto, Justin and Sepah, Naz and Raff, Edward and Madan, Kanika and Voleti, Vikram and Ebrahimi Kahou, Samira and Michalski, Vincent and Serdyuk, Dmitriy and Arbel, Tal and Pal, Chris and Varoquaux, Ga{\"e}l and Vincent, Pascal},
  title     = {Accounting for Variance in Machine Learning Benchmarks},
  booktitle = {Proceedings of Machine Learning and Systems (MLSys)},
  year      = {2021}
}

@article{hooker2021unrestricted,
  author  = {Hooker, Giles and Mentch, Lucas and Zhou, Siyu},
  title   = {Unrestricted Permutation Forces Extrapolation: Variable Importance Requires at Least One More Model, or There Is No Free Variable Importance},
  journal = {Statistics and Computing},
  volume  = {31}, number = {6}, pages = {82}, year = {2021}
}

@article{mitchell2022,
  author  = {Mitchell, Rory and Cooper, Joshua and Frank, Eibe and Holmes, Geoffrey},
  title   = {Sampling Permutations for {Shapley} Value Estimation},
  journal = {Journal of Machine Learning Research},
  volume  = {23}, number = {43}, pages = {1--46}, year = {2022}
}

@article{damour2022,
  author  = {D'Amour, Alexander and Heller, Katherine and Moldovan, Dan and Adlam, Ben and Alipanahi, Babak and Beutel, Alex and Chen, Christina and Deaton, Jonathan and Eisenstein, Jacob and Hoffman, Matthew D. and others},
  title   = {Underspecification Presents Challenges for Credibility in Modern Machine Learning},
  journal = {Journal of Machine Learning Research},
  volume  = {23}, number = {226}, pages = {1--61}, year = {2022}
}

@article{fisher2019,
  author  = {Fisher, Aaron and Rudin, Cynthia and Dominici, Francesca},
  title   = {All Models are Wrong, but Many are Useful: Learning a Variable's Importance by Studying an Entire Class of Prediction Models Simultaneously},
  journal = {Journal of Machine Learning Research},
  volume  = {20}, number = {177}, pages = {1--81}, year = {2019}
}

@article{dong2020,
  author  = {Dong, Jiayun and Rudin, Cynthia},
  title   = {Exploring the Cloud of Variable Importance for the Set of All Good Models},
  journal = {Nature Machine Intelligence},
  volume  = {2}, number = {12}, pages = {810--824}, year = {2020}
}

@article{krishna2024,
  author  = {Krishna, Satyapriya and Han, Tessa and Gu, Alex and Wu, Steven and Jabbari, Shahin and Lakkaraju, Himabindu},
  title   = {The Disagreement Problem in Explainable Machine Learning: A Practitioner's Perspective},
  journal = {Transactions on Machine Learning Research},
  year    = {2024}
}

@article{guyon2003,
  author  = {Guyon, Isabelle and Elisseeff, Andr{\'e}},
  title   = {An Introduction to Variable and Feature Selection},
  journal = {Journal of Machine Learning Research},
  volume  = {3}, pages = {1157--1182}, year = {2003}
}

@article{strobl2008,
  author  = {Strobl, Carolin and Boulesteix, Anne-Laure and Kneib, Thomas and Augustin, Thomas and Zeileis, Achim},
  title   = {Conditional Variable Importance for Random Forests},
  journal = {BMC Bioinformatics},
  volume  = {9}, pages = {307}, year = {2008}
}

@article{peng2005,
  author  = {Peng, Hanchuan and Long, Fuhui and Ding, Chris},
  title   = {Feature Selection Based on Mutual Information: Criteria of Max-Dependency, Max-Relevance, and Min-Redundancy},
  journal = {IEEE Transactions on Pattern Analysis and Machine Intelligence},
  volume  = {27}, number = {8}, pages = {1226--1238}, year = {2005}
}
}
\end{document}